\pdfoutput=1

\documentclass[10pt,twocolumn,letterpaper]{article}

\usepackage[pagenumbers]{style}

\usepackage{array}
\usepackage{svg}
\usepackage{graphicx}
\usepackage{amsmath}
\usepackage{amssymb}
\usepackage{booktabs}
\usepackage{multirow}
\usepackage{xspace}
\usepackage{xcolor}
\usepackage{svg}
\usepackage{tikz}
\usepackage[percent]{overpic}
\usepackage{rotating}
\usepackage{scalerel}
\usepackage{bm}

\usepackage[pagebackref,breaklinks,colorlinks]{hyperref}

\usepackage[capitalize]{cleveref}
\crefname{section}{Sec.}{Secs.}
\Crefname{section}{Section}{Sections}
\Crefname{table}{Table}{Tables}
\crefname{table}{Tab.}{Tabs.}

\begin{document}

\title{VHS: High-Resolution Iterative Stereo Matching with Visual Hull Priors}

\author{
  \begin{minipage}{3.75cm} \centering Markus Plack \end{minipage}
  \begin{minipage}{3.75cm} \centering Hannah Dr{\"o}ge \end{minipage}
  \begin{minipage}{3.75cm} \centering Leif Van Holland \end{minipage}
  \begin{minipage}{3.75cm} \centering Matthias B. Hullin \end{minipage}
  \\
University of Bonn \\
Bonn, Germany \\
{\tt\small \{mplack,droege,holland,hullin\}@cs.uni-bonn.de}
}
\twocolumn[{%
\renewcommand\twocolumn[1][]{#1}%
\maketitle
\graphicspath{{figures/teaser/}}
{
  \def\svgwidth{0.99\linewidth}
  \footnotesize
  \begingroup%
  \makeatletter%
  \providecommand\color[2][]{%
    \errmessage{(Inkscape) Color is used for the text in Inkscape, but the package 'color.sty' is not loaded}%
    \renewcommand\color[2][]{}%
  }%
  \providecommand\transparent[1]{%
    \errmessage{(Inkscape) Transparency is used (non-zero) for the text in Inkscape, but the package 'transparent.sty' is not loaded}%
    \renewcommand\transparent[1]{}%
  }%
  \providecommand\rotatebox[2]{#2}%
  \newcommand*\fsize{\dimexpr\f@size pt\relax}%
  \newcommand*\lineheight[1]{\fontsize{\fsize}{#1\fsize}\selectfont}%
  \ifx\svgwidth\undefined%
    \setlength{\unitlength}{989.99997837bp}%
    \ifx\svgscale\undefined%
      \relax%
    \else%
      \setlength{\unitlength}{\unitlength * \real{\svgscale}}%
    \fi%
  \else%
    \setlength{\unitlength}{\svgwidth}%
  \fi%
  \global\let\svgwidth\undefined%
  \global\let\svgscale\undefined%
  \makeatother%
  \begin{picture}(1,0.36363637)%
    \lineheight{1}%
    \setlength\tabcolsep{0pt}%
    \put(0,0){\includegraphics[width=\unitlength]{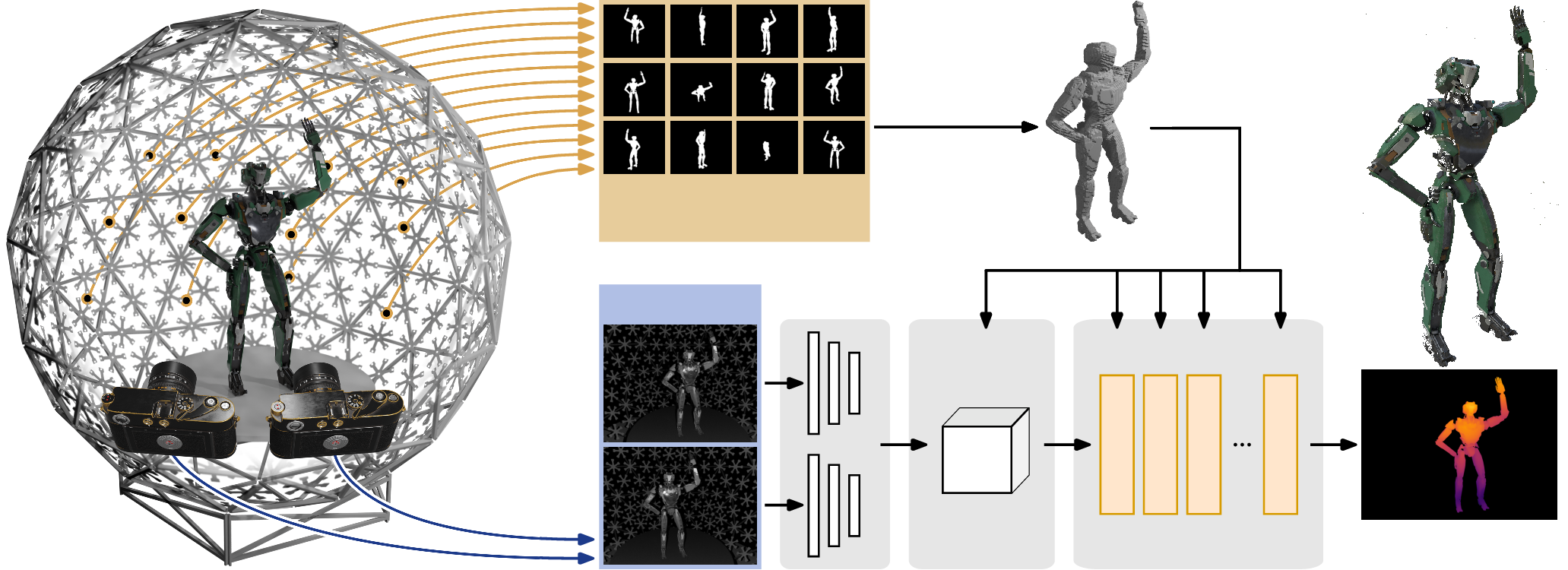}}%
    \put(0.46855537,0.23488934){\color[rgb]{0,0,0}\makebox(0,0)[t]{\lineheight{1.25}\smash{\begin{tabular}[t]{c}Masks from\\Auxiliary Views\end{tabular}}}}%
    \put(0.43189963,0.1637386){\color[rgb]{0,0,0}\makebox(0,0)[t]{\lineheight{1.25}\smash{\begin{tabular}[t]{c}Stereo Pair\end{tabular}}}}%
    \put(0.62666496,0.01139904){\color[rgb]{0,0,0}\makebox(0,0)[t]{\lineheight{1.25}\smash{\begin{tabular}[t]{c}Sparse Init.\end{tabular}}}}%
    \put(0.76629156,0.13756002){\color[rgb]{0,0,0}\makebox(0,0)[t]{\lineheight{1.25}\smash{\begin{tabular}[t]{c}Weak Prior\end{tabular}}}}%
    \put(0.60645219,0.28939753){\color[rgb]{0,0,0}\makebox(0,0)[t]{\lineheight{1.25}\smash{\begin{tabular}[t]{c}Visual Hull\end{tabular}}}}%
    \put(0.62856762,0.13780272){\color[rgb]{0,0,0}\makebox(0,0)[t]{\lineheight{1.25}\smash{\begin{tabular}[t]{c}Strong Prior\end{tabular}}}}%
    \put(0.9309077,0.01150222){\color[rgb]{0,0,0}\makebox(0,0)[t]{\lineheight{1.25}\smash{\begin{tabular}[t]{c}Disparity Map\end{tabular}}}}%
    \put(0.59876543,0.09732317){\color[rgb]{0,0,0}\makebox(0,0)[t]{\lineheight{1.25}\smash{\begin{tabular}[t]{c}k\end{tabular}}}}%
    \put(0.5954032,0.06700122){\color[rgb]{0,0,0}\makebox(0,0)[t]{\lineheight{1.25}\smash{\begin{tabular}[t]{c}h\end{tabular}}}}%
    \put(0.6262797,0.03797614){\color[rgb]{0,0,0}\makebox(0,0)[t]{\lineheight{1.25}\smash{\begin{tabular}[t]{c}w\end{tabular}}}}%
    \put(0.76128352,0.01139904){\color[rgb]{0,0,0}\makebox(0,0)[t]{\lineheight{1.25}\smash{\begin{tabular}[t]{c}Dense Refinement\end{tabular}}}}%
    \put(0.76384036,0.28939753){\color[rgb]{0,0,0}\makebox(0,0)[t]{\lineheight{1.25}\smash{\begin{tabular}[t]{c}Limits\end{tabular}}}}%
  \end{picture}%
\endgroup%

}

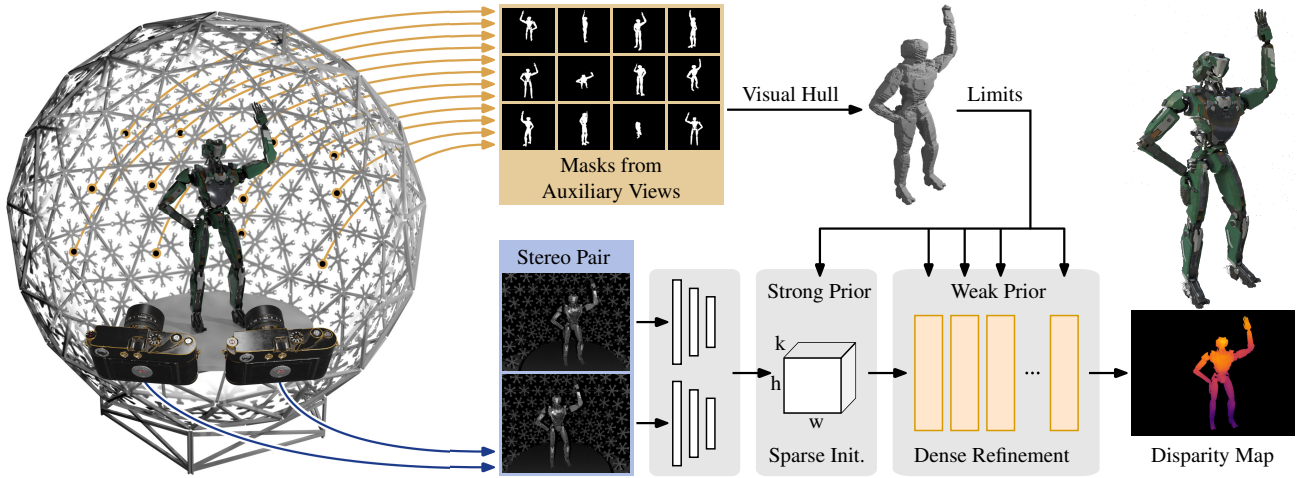
\captionof{figure}{
  We propose a technique to induce a rough shape estimate from object masks (top) as prior information to a
  novel, sparse-dense stereo-matching network (bottom) for the application in capture stages (left)
  for accurate and memory-efficient disparity estimation (right).
}
\label{fig:teaser}
\vspace{8pt}

}]

\begin{abstract}
We present a stereo-matching method for depth estimation from high-resolution images using visual hulls as priors, and a memory-efficient technique for the correlation computation. Our method uses object masks extracted from supplementary views of the scene to guide the disparity estimation, effectively reducing the search space for matches. This approach is specifically tailored to stereo rigs in volumetric capture systems, where an accurate depth plays a key role in the downstream reconstruction task. To enable training and regression at high resolutions targeted by recent systems, our approach extends a sparse correlation computation into a hybrid sparse-dense scheme suitable for application in leading recurrent network architectures.

We evaluate the performance-efficiency trade-off of our method compared to state-of-the-art methods, and demonstrate the efficacy of the visual hull guidance. In addition, we propose a training scheme for a further reduction of memory requirements during optimization, facilitating training on high-resolution data.
\end{abstract}

\section{Introduction}
\label{sec:intro}

Stereo matching is a long-standing problem in the area of computer vision, driving core functionality in a wide range of applications, for example in the automotive industry, virtual and augmented reality systems, as well as in medical imaging, agriculture, remote sensing, and robotics domains.
Recently, interest surged in telepresence and virtual production scenarios that use volumetric capturing systems~\cite{guo2019relightables, collet2015high,orts2016holoportation,heagerty2024holocamera}, which rely on fast and accurate depth estimates for downstream reconstruction tasks.
The disparity regression problem is typically solved by initially computing the matching cost between a stereo image pair or a suitable feature representation thereof and searching for the best correspondences along the epipolar lines resulting in a highly irregular cost landscape.
Challenges include occlusion, view-dependent reflectivity, repetitive patterns, and insufficient calibration accuracy.
With the rise of deep learning in the domain of computer vision, classical matching methods \cite{hamzah2016literature, barnard1980disparity, scharstein2003high, muhlmann2002calculating} are surpassed by data-driven approaches \cite{mayer2016large, kendall2017end, xu2020aanet, guo2019group}. 
Recently, so-called all-pairs-correlation networks based on the optical flow network RAFT~\cite{teed2020raft} have shown to perform remarkably well when applied in the stereo matching context~\cite{lipson2021raft}.
Those methods compute a dense correlation volume for \emph{all} possible matches and perform stereo regression in an iterative fashion akin to gradient descent methods.
One distinct drawback of such approaches is that the size of the full correlation volume scales quadratically with the horizontal input resolution, limiting their applicability on high-resolution inputs.
One solution to reduce the prohibitive memory requirement is to use sparse representations~\cite{wang2021scv} that only store the $k$ most relevant entries of the correlation volume, similar to $k$-nearest-neighbor ($k$NN) methods.
While this still requires the computation of \emph{all} correlation values, %
which does not reduce the computational costs, the memory demand only scale linearly with respect to the horizontal input resolution,
but possibly discards valuable information.

In contrast, we propose a sparse-dense approach that allows us to consider all disparities, avoiding the limitations associated with missing values in sparse representations. We calculate disparities using a sparse method initially, followed by a refinement in a memory-efficient dense manner.
As a crucial step to reduce the amount of sparse candidates, we propose to employ the visual hull \cite{laurentini1994visual} as a rough shape estimate that reduces the set of valid disparities to points inside the hull. The foreground segmentation masks required for this are available through the use of chroma-keying \cite{raditya2021effectivity} or more sophisticated image-level segmentation approaches \cite{guo2019relightables} in many capturing scenarios and thus the visual hull can be computed easily. During the refinement step, we can further use the hull as a weak prior.

In summary, our contributions are as follows:
\begin{itemize}
  \item We present a method to induce prior knowledge of visual hulls from auxiliary views into a recurrent stereo-matching network to reduce the initial disparity search space and as guidance for the iterative refinement.
  \item We demonstrate a sparse-dense correlation method that effectively reduces peak memory requirements while retaining the accuracy of all-pairs correlation methods through just-in-time computation for the updates.
  \item We propose an optimization scheme to realize high-resolution training of recurrent stereo network architectures
  and show how the visual hull-guided network can benefit from pre-training on conventional training data by making the input optional.
\end{itemize}
We share the model and training implementation of our \textbf{V}isual \textbf{H}ull \textbf{S}tereo (\textit{VHS}) network and the custom kernels along with the data used for training and testing
at \url{https://github.com/unlikelymaths/vhs}. %

\begin{figure*}[htb]
\centering

\scriptsize
\begin{overpic}[width=1.0\linewidth]
{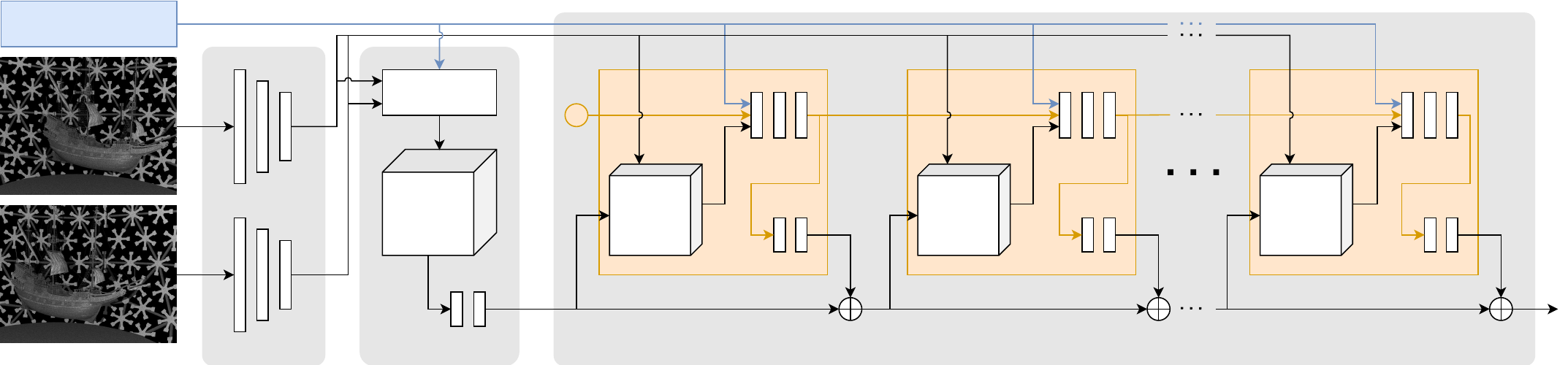}

\put(0.8, 21.4){Visual Hull Prior}

\put(24.5, 17.5){VH filtering}
\put(25.6, 16.3){+ $k$NN}
\put(24, 13){$k$}
\put(24.6, 9.5){$h$}
\put(26.5, 7.3){$w$}

\put(29.5, 5.2){$g$}
\put(31, 1.9){$D^0$}

\put(35.6, 17.2){$H^0$}
\put(38.4, 12.8){$k'$}
\put(39.1, 9.1){$h$}
\put(40.8, 7.3){$w$}
\put(54.4, 5.2){$\Delta$}
\put(54.7, 2.0){$D^1$}

\put(58.0, 12.8){$k'$}
\put(58.7, 9.1){$h$}
\put(60.4, 7.3){$w$}
\put(74.0, 5.2){$\Delta$}
\put(74.5, 2.0){$D^2$}

\put(79.8, 12.8){$k'$}
\put(80.5, 9.1){$h$}
\put(82.2, 7.3){$w$}
\put(95.8, 5.2){$\Delta$}
\put(95.7, 1.7){$D^n$}

\put(0.7, 0.3){Stereo Image Pair}
\put(13.3, 0.4){Feature Ext.}
\put(23.5, 0.4){Initial Disparity}
\put(56, 0.4){Iterative Refinement with ConvGRU}

\end{overpic}

\caption{
  Overview of the three stages of our disparity estimation network VHS.
  Following the \emph{Feature Extraction} we compute an \emph{Initial Disparity} estimate $D_0$ from a sparse $k$NN cost volume restricted by the visual hull.
  Next, we perform an \emph{Iterative Refinement} of the disparity guided by the visual hull prior using ConvGRU modules and dense local correlations with window size $k'$.
}
\label{fig:overview}
\end{figure*}
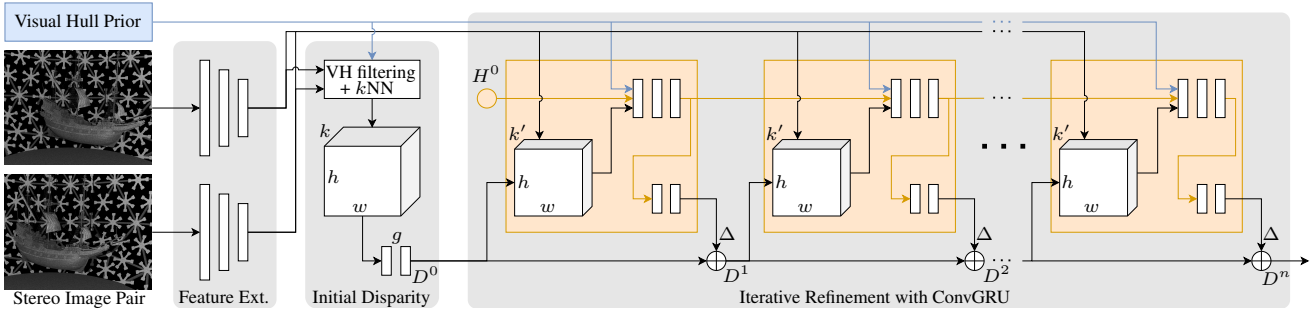

\section{Related Work}
\label{sec:related}

Learning-based methods using correlation volumes to predict accurate disparity maps have shown great potential in stereo matching. We briefly review approaches for generating cost volumes and discuss previous work on further refinement of the disparities by iterative update methods before giving an overview of stereo vision approaches targeting efficiency aspects. 

\subsection{Matching Cost Volume} \label{sec:rel_cost_volumes}

Recent developments in end-to-end learning approaches for cost volumes have successfully captured the similarity of pixel pairs across varied degrees of disparity in stereo matching \cite{zhang2019ga, kendall2017end, mayer2016large, gu2020cascade}.

In this context, Mayer \etal \cite{mayer2016large} introduced a method based on \textit{correlation} for calculating cost volume, followed by subsequent work  \cite{liang2018learning, tonioni2019real}. This approach measures the correlation between the features of two images within a 1D correlation layer applied horizontally along the disparity line. 

\textit{Concatenation}-based methods \cite{ li2022practical, nie2019multi, chabra2019stereodrnet, abd2024refining}, on the other hand, follow a different strategy. 
Kendall \etal \cite{kendall2017end} concatenated unary features with their corresponding features along the disparity line. They generated a 4D cost volume, subsequently processed through an encode-decoder network with 3D convolutions across spatial dimensions and disparity. 
To further regularize the 4D cost volume,  Chang \etal \cite{chang2018pyramid} discussed the implementation of a learned regularization using a stacked hourglass network. 
Addressing the lack of explicit similarity measures in previous {concatenation}-based approaches, Guo \etal~\cite{guo2019group} proposed integrating group-wise correlations into the 4D cost volume by dividing features into sub-groups and calculating correlations for each. 
To improve the performance even in regions with less texture, recent work \cite{xu2022attention} filters the concatenation volume with
attention weights to suppress unnecessary information.

To overcome storage and runtime limitations, \textit{cascading}  cost volumes were created by building a cost volume pyramid and progressively refining depth estimation with a coarse-to-fine technique \cite{gu2020cascade}. 
Other cascade formulations have been proposed for even higher resolutions \cite{wang2021patchmatchnet} or address unbalanced disparity distributions \cite{shen2021cfnet}.%

\subsection{Iterative Updates in Stereo Matching}

Initially proposed for optical flow estimation, deep learning approaches have successfully employed traditional optimization methods using learned updates to improve performance.
These methods refine disparity maps through successive updates, as 
demonstrated by RAFT (Recurrent All-Pairs Field Transforms)~\cite{teed2020raft}. 
RAFT consists of a feature encoding step, computation of correlation volumes containing the correlations between all pixel pairs, and a learned update operator that iteratively updates the optical flow estimation based on the correlation volumes.  
Based on this, Lipson \etal~\cite{lipson2021raft} introduced an adaptation of RAFT for stereo disparity estimation, called RAFT-Stereo, which recurrently updates the disparity map using local cost values.  

Several works introduced modifications to this idea.
IGEV-Stereo~\cite{xu2023iterative} introduces the geometry encoding volume to extend the all-pairs correlation volume and regress a better initial disparity. 
Instead of using the GRU to update the flow field, Wang \etal~\cite{wang2022itermvs} repurposed it to predict the depth probability of each pixel. Zhao \etal~\cite{zhao2023high} propose improvements in the iterative process to preserve detail in the hidden state by decoupling the disparity map from the hidden state and implementing a normalization strategy to handle large variations in disparities. 
EAI-Stereo~\cite{zhao2022eai} replaced the GRU with an error-aware iterative module.

\subsection{Efficiency}

In a structured light setting~\cite{le1988structured,vuylsteke1990range,martinez2013kinect},
 projected patterns are designed to uniquely identify the depth of objects at each position. 
Hence, the problem can be solved more efficiently for known light patterns,
as demonstrated by \eg Hyperdepth~\cite{fanello2016hyperdepth}
using a random forest approach and 
the branching network in Gigadepth~\cite{schreiberhuber2022gigadepth}.
Note that this is different from our setting based on the work of Guo \etal~\cite{guo2019relightables}
where multiple, potentially overlapping, patterns are projected into the scene.

Turning to wider stereo vision challenges, the bottleneck with cost volumes is their large search space, which requires considerable computation and storage to find the desired disparity.
Khamis~\etal~\cite{khamis2018stereonet} reduced the computational cost by refining the disparity from a low-resolution cost volume through multiple levels of resolution.
Additionally, recent works~\cite{bangunharcana2021correlate,wang2021fadnet++} stress  real-time disparity estimation in stereo vision. 
While Shamsafar \etal~\cite{shamsafar2022mobilestereonet} relies on lightweight architectures to optimize resources, 
Garrepalli~\etal introduced DIFT~\cite{garrepalli2023dift} as a mobile architecture for optical flow that uses
just-in-time computation of the correlation to reduce peak memory use and served as the inspiration for our correlation computation in the iterative updates.
SCV-Net~\cite{lu2018sparse} builds a sparse correlation volume that resembles dilated convolutions controlled via a fixed sparsity value and without dependence on the inputs.
Lastly, SCV-Stereo~\cite{wang2021scv} is an alternative approach to sparse correlation volumes.
Different from their method, we use $k$NN correlation for the initial disparity estimate instead of zero initialization and compute dense correlations on an ad hoc basis during the iterative stages.

\section{Visual Hull Stereo}

The overall structure of our method is based on RAFT-Stereo~\cite{lipson2021raft}
and is shown in~\cref{fig:overview}. 
It consists of three stages. First, the pair of input images is encoded into a feature representation using a pre-trained encoding network.
These features are then used to compute an initial correlation cost volume. Together with prior information attained from a set of image masks of the scene, a sparse set of $k$ disparities with the highest correlation values is selected from which an initial disparity value is estimated (\cref{sec:sparse_correlation,sec:visual_hull_prior}).
Following, the disparity is iteratively refined using a \textit{Convolutional Gated Recurrent Unit} (ConvGRU)-based network and upsampling network~\cite{xu2023iterative}, without the need to hold the full cost volume in memory at any time (\cref{sec:iterative_optimization}).

\subsection{Sparse Correlation}
\label{sec:sparse_correlation}

Given a rectified stereo pair, we use a shared feature encoding network~\cite{xu2023iterative} to extract features at 25\% of the original image size.
This representation is used to compute an initial set of the $k$ best matches. First, we define the cost $c_p(d)\in\mathbb{R}$ of disparity $d\in [0, w]$ at pixel $p\in \mathbb{N}^2$ as the inner product of the corresponding feature vectors {$f_p, g_{p-(0,d)^T}$}, from the left and right pictures of size $h\times w$, where $g_{p-(0,d)^T}$ represents the feature vector at the pixel in the right image offset by $d$:
\begin{equation}
    c_p(d) = f_p \cdot g_{p-(0,d)^T}
    \label{eq:initial_costs}
\end{equation}

Storing the full set of correlation values at high resolutions can be inefficient and resource-intensive, as the dense cost volume scales quadratically with the image width when the maximal disparity is properly adjusted. %
To decrease the memory requirements, we instead use a sparse correlation cost volume, which assigns to each pixel $p$ a much smaller subset of correlation values $c$ and corresponding disparity values $d$,
\begin{equation}
\mathcal{M}_{p} = \{(d, c_p(d)) \,|\, d \in \mathcal{D}^{k\text{NN}}_p\},
\end{equation}
where $\mathcal{D}^{k\text{NN}}_p$ represents the set of $k$ best disparities for each pixel:
\begin{equation}\label{eq:knn}
    \mathcal{D}^{k\text{NN}}_p = \underset{\tilde{\mathcal{D}}_p \subset \mathcal{D}_p, |\tilde{\mathcal{D}}_p| = K}{\arg\max} \sum_{d \in \tilde{\mathcal{D}}_p} c_p(d)
\end{equation}
Here, $\mathcal{D}_p$ is the set of all disparity candidates for pixel $p$.

\subsection{Visual Hull Prior}
\label{sec:visual_hull_prior}

\begin{figure}[t]
    \centering
    \input{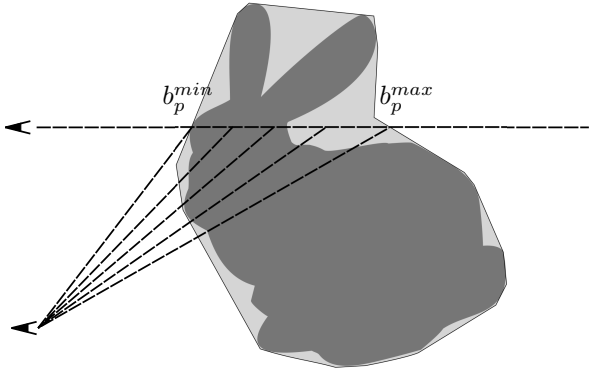}
    \caption{Estimation of the disparity boundaries $(b_p^{min}, b_p^{max})$, from two rectified views of an object's visual hull. The visual hull encloses the objects' surface, so the surface is guaranteed to lie within the disparity boundaries.} 
    \label{fig:visualhull_minmax}
\end{figure}

This search for the best candidates can be further improved by inducing a prior based on image masks from the scene. The visual hull, as defined in \cite{laurentini1994visual}, provides an efficient approximation of an object's shape derived from silhouettes captured by multiple cameras.
In adherence to the representation proposed in \cite{scharr2017fast}, we compute the visual hull using a collection of masked input images, which is stored within an octree structure for compact storage and fast access. The octree is designed such that each leaf node 
indicates whether it is inside or outside the visual hull.
Given this information, we calculate the hull boundaries by sampling rays projected into the scene from the reference view and evaluating these rays for transitions between outside and inside regions of objects.
From these transitions, we create depth limits for each camera viewpoint and define disparity boundaries $b_p=(b_p^{min}, b_p^{max})$ based on pixel location $p$, as illustrated in Figure \ref{fig:visualhull_minmax}. 
The insight that the surfaces of the objects are confined within the interval $[b_p^{\min}, b_p^{\max}]$ can be leveraged to reduce computational requirements when computing the initial disparity map $D^0$.

We streamline the $k$-nearest-neighbor search, previously performed across an expansive set of disparity candidates $\mathcal{D}_p$ for pixel $p$ as described in (\ref{eq:knn}), by focusing only on disparities constrained within $b_p$:
\begin{equation}
    {\mathcal{D}}^*_p = \{d\,|\, b_p^{min} \leq d \leq b_p^{max}\}, \qquad {\mathcal{D}}^*_p \subseteq \mathcal{D}_p
\end{equation}
This approach allows for a faster computation of the restricted correlation cost volume ${\mathcal{M}_p^*}$ by skipping unnecessary evaluations of the correlation. Accordingly, we define our initial disparity map as follows:
\begin{equation}
    D^0_{p} = \sum_{l=1}^K d_{l} \cdot g(c_p(d))_{l}, \qquad  (d, c_p(d)) \in {\mathcal{M}}_p^*
\end{equation}
where $g$ is an attention-based transformation network with a softmax function as the last layer.

\subsection{Iterative Disparity Refinement}
\label{sec:iterative_optimization}

We use a hierarchical ConvGRU network on three resolutions to iteratively refine the predicted disparities starting with the initial values $D_p^0$, similar to \cite{xu2023iterative}:
The network updates a hidden state $H^i$ taking the current disparity values and contextual features extracted from the corresponding image data, and the correlated features around the current disparity estimate as input.
The new state is used to predict an offset~$\Delta_p^i$ from which the refined disparity values are computed as 
\begin{equation}
   D_p^{i+1} = D_p^i + \Delta_p^i. 
\end{equation}
\paragraph{Memory Efficient Correlation} Instead of sampling correlation values from a pre-computed full cost volume, we propose to compute a local correlation volume ad hoc to reduce memory usage. 
This volume is bounded within a window $W_p^i$ of size \mbox{$2r+1$}, which is centered on the currently estimated disparity $D_p^i$,
\begin{equation}
    W_p^{i} = [D^i_p - r, D^i_p + r],
\end{equation}
where we fix $r = 4$ following~\cite{xu2023iterative}.
We compute the correlations group-wise, as originally proposed by~\cite{guo2019group}, by dividing the feature vectors %
into a set of subgroups. 
Please note that, for the initial disparity $D_p^0$, we  strategically omitted the group-wise correlation calculation. This is due to the complexity of uniquely defining $k$NN for group-wise correlations, ensuring that our approach remains computationally efficient.

\paragraph{Visual Hull as Weak Prior} As additional information, we supply the ConvGRU with a flag $f_p(d)$ that guides the network to predict a value within the visual hull,
\begin{equation}
f_p(d) = \begin{cases} 
 1 & \text{if } d \in {D}^*_p, \\
-1 & \text{otherwise}
\end{cases}
\end{equation}
for each disparity value $d$ within the window $W_p^{i}$. 
In that way, the limits $b_p$ obtained from the visual hull operate as a weak prior
guiding the disparity regression while retaining valuable correlation information
for cases such as incorrect limits due to masking errors.

One distinct advantage of our visual hull guidance is that the disparity limits are an optional input to the whole pipeline.
During the initial sparse correlation, we can fall back to sampling from all values below a pre-defined threshold in the same manner as established models, and during the dense updates, we set $f_p(d) = 0$ to indicate missing information.
This enables the application of our sparse correlation method even without masked measurements and pre-training of our method on existing datasets.

\section{Training Details}
Given the particular nature of our method in terms of target application and required inputs, a boilerplate training procedure following the literature would be unproductive.
Therefore, we present custom training details tailored to our use case, covering the preparation of custom data along with training strategies.
We further introduce a memory-efficient approach enabling training at even higher resolutions. 

\subsection{Dataset Preparation}\label{sec:dataset_preparation}
\begin{figure}[tb]

\begin{minipage}[b]{0.49\linewidth}
  \centering
  \centerline{\includegraphics[width=\linewidth]{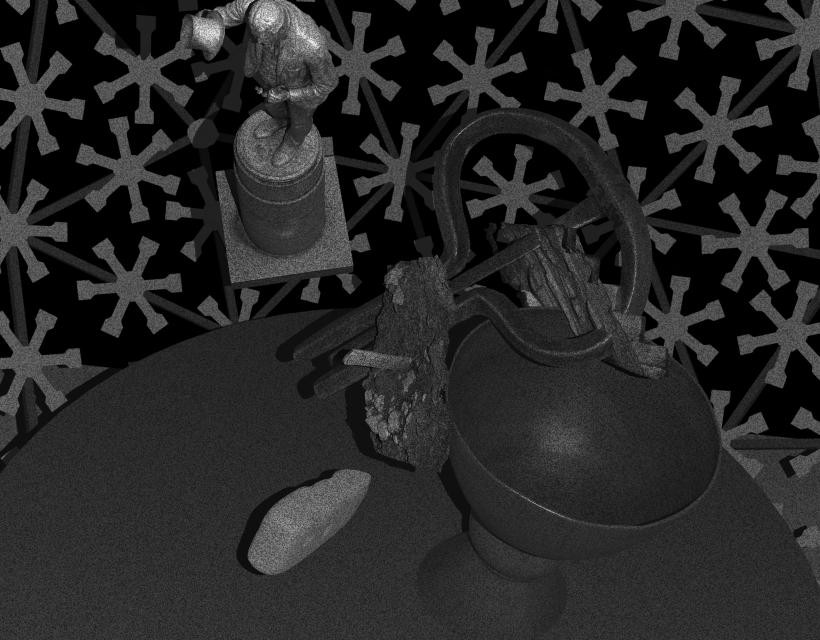}}
  \centerline{\small (a) Reference Image}\medskip
\end{minipage}
\hfill
\begin{minipage}[b]{0.49\linewidth}
  \centering
  \centerline{\includegraphics[width=\linewidth]{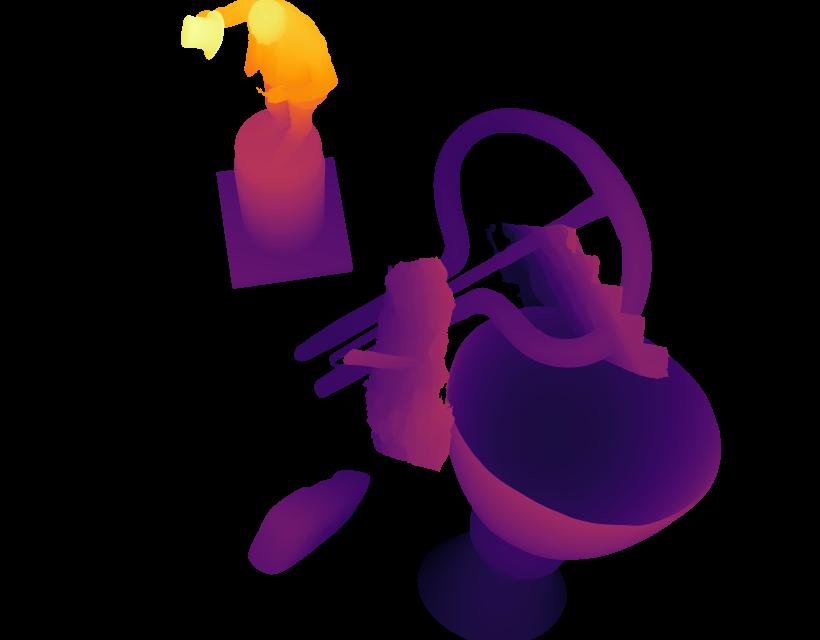}}
  \centerline{\small (b) Ground Truth Disparity}\medskip
\end{minipage}

\begin{minipage}[b]{0.49\linewidth}
  \centering
  \centerline{\includegraphics[width=\linewidth]{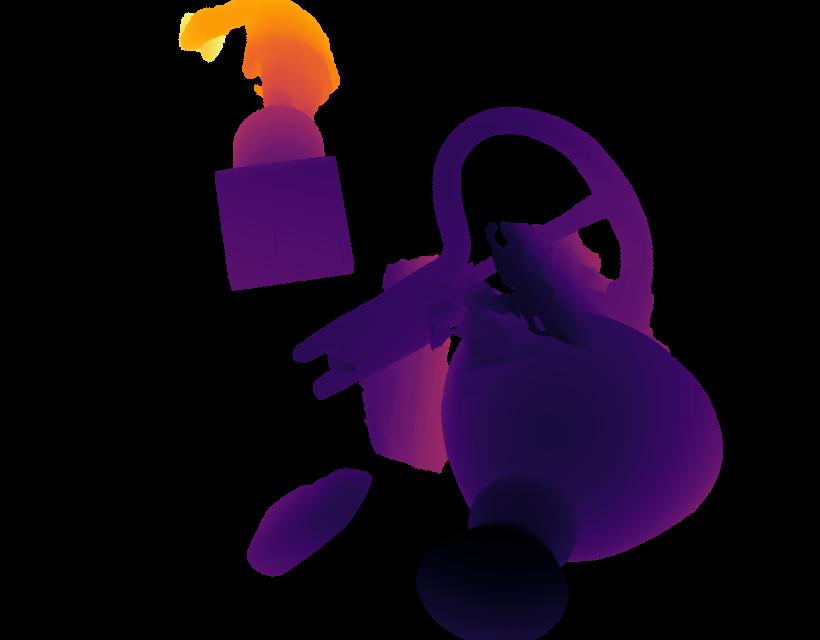}}
  \centerline{\small (c) Disparity Limit $b^{min}$}
\end{minipage}
\hfill
\begin{minipage}[b]{0.49\linewidth}
  \centering
  \centerline{\includegraphics[width=\linewidth]{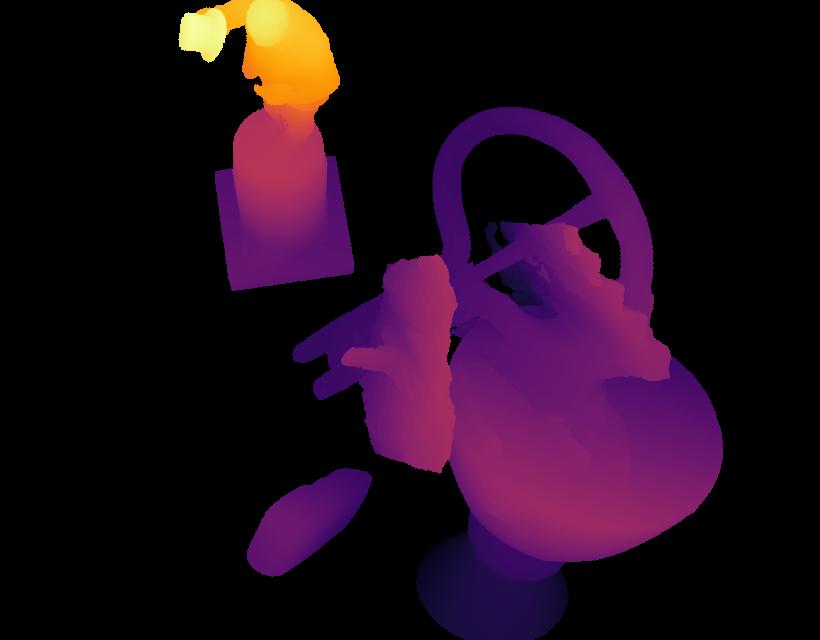}}
  \centerline{\small (d) Disparity Limit $b^{max}$}
\end{minipage}

\caption{
  Sample from the FlyingObjaverse training dataset.
  Notice how the true disparity is close to the upper disparity limit
  except for the basin in the bottom right, which cannot be recovered from the visual hull.
}
\label{fig:flying_dataset_examples}
\end{figure}

Common stereo datasets like SceneFlow~\cite{mayer2016large} do not contain ground truth meshes or auxiliary views, which prevents the extraction of a meaningful visual hull. 
As an alternative, we render a custom dataset with Mitsuba 3~\cite{jakob2022mitsuba} and meshes from Objaverse-XL~\cite{deitke2023objaverse} to train our network. 
The dataset generation loosely follows the approach of SceneFlow by placing objects on a virtual capture stage.
Each scene contains a randomly transformed arrangement of $1-10$ objects, as shown in \cref{fig:flying_dataset_examples}, with an infrared camera stereo setup using active illumination with projected patterns similar to~\cite{guo2019relightables}
and a total of $68$ cameras for the masks, all captured at a resolution of $4608 \times 5328$.
We render $2$ stereo pairs for $500$ scenes.

For testing, we follow the same rendering pipeline but select meshes from different sources to avoid contamination of the training dataset.
To test performance on difficult lighting effects, we curated scenes with objects that include challenging reflectance properties and fine details using high-quality meshes from Polyhaven\footnote{\url{https://polyhaven.com/}} and build eight scenes, each viewed from four different angles.
As a second test set, we used SMPL \cite{loper2015smpl} human models with texture from SMPLitex \cite{casas2023smplitex} to evaluate performance on human subjects.
We create $100$ scenes by combining random poses from the animations with random textures and render $2$ stereo pairs for each scene.

\subsection{Training Strategy}

Having the visual hull guidance as an entirely optional component, allows our method to harness a more flexible training process and to predict the disparity map even without any pre-calculated masks.
We use this flexibility in our experiments by pre-training a base model on Sceneflow \cite{mayer2016large} and subsequently fine-tuning the network on our custom training data.
The training is performed on SceneFlow final pass for $20$ epochs using AdamW~\cite{loshchilov2017decoupled} with a one-cycle learning rate schedule with a learning rate of $0.00015$ and a batch size of $4$.
We use random crops of size $288 \times 640$, random y-jitter and occlusion as augmentation, and an $L_1$ loss following the weighting of RAFT-Stereo~\cite{lipson2021raft}.
This model serves as our baseline for a benchmark evaluation on the SceneFlow test set.
Subsequently, the network is fine-tuned on the simulated data of Objaverse-XL (\cref{sec:dataset_preparation}) for high-resolution stereo following the same settings, except for a magnified random cropping of $256 \times 2048$, batch size of $1$ and with the additional visual hull inputs, which we randomly drop for $\frac{1}{8}$ of the samples.
Note that we use RGB inputs for the benchmark comparison and greyscale for the simulation of IR images for all other experiments.

\paragraph{Memory Efficient Training}

\begin{figure}[htb]

\begin{overpic}[width=1.0\linewidth]
{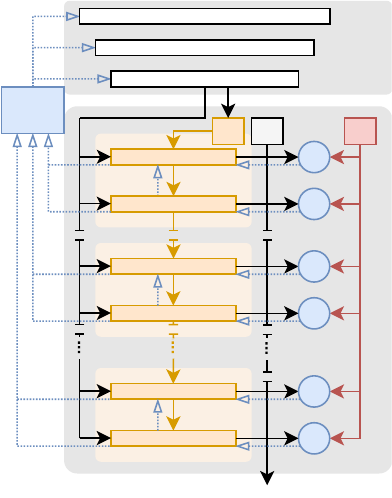}
\put(3, 77){$\sum \Delta$}
\put(43.6, 72.7){$H_0$}
\put(51.5, 72.7){$D_0$}
\put(69.8, 72.7){$D_{\text{gt}}$}
\put(61.2, 67.5){$\mathcal{L}_1$}
\put(61.2, 58){$\mathcal{L}_2$}
\put(61.2, 45.5){$\mathcal{L}_3$}
\put(61.2, 36.1){$\mathcal{L}_4$}
\put(60.4, 20.5){$\mathcal{L}_{\scaleto{n-1}{3pt}}$}
\put(61.2, 11){$\mathcal{L}_{n}$}
\put(73, 25){
    \begin{turn}{90}
        Iterative Refinement
    \end{turn}
}

\put(70, 85){
    \begin{turn}{90}
        Feature
    \end{turn}
}
\put(74, 83.5){
    \begin{turn}{90}
        Encoding
    \end{turn}
}

\end{overpic}
\centering

\caption{
  Memory efficient training scheme for $n=2$ consecutive update steps.
  After the computation of the losses $\mathcal{L}_i$ and $\mathcal{L}_{i+1}$, we perform backpropagation to accumulate gradients of the update network parameters and detach the hidden state effectively freeing the computational graph.
  $\sum \Delta$ indicates an optional accumulation of gradients to avoid multiple backward passes through the feature extraction network.
}
\label{fig:memory_efficient_backprop}
\end{figure}
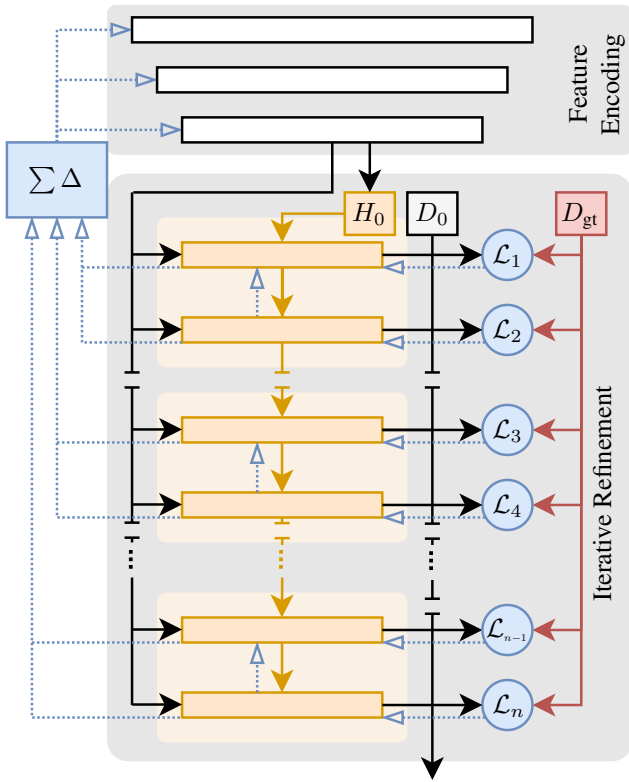

During the training of most iterative methods, each update of the disparity consumes more VRAM since the full compute graph needs to be stored in memory.
We propose to split the forward and backward computation in a manner that reduces the memory requirement while still retaining accurate gradient information as shown in \cref{fig:memory_efficient_backprop}.
For $n$ consecutive update steps we compute the losses on the upscaled disparity predictions as usual.
Then, we backpropagate the partial loss and detach the hidden state such that the computational graph can be erased.
To avoid multiple backward passes through the costly feature extraction network, we propose to optionally accumulate all gradients for the feature vectors first before performing a final backpropagation after all iterations are through.

\paragraph{Technical Details}
Using CUDA, we build a visual hull octree from rendered masks from which the disparity limits are computed. 
Our network is implemented in Pytorch with custom CUDA kernels for the correlation computations and we use warp-level shuffle operations to make the initial $k$NN correlation computation efficient. 
As such, the number of candidates is limited to $32$, but we use $8$ for all experiments following~\cite{wang2021scv}. All our experiments were conducted on an {NVIDIA GeForce RTX 4090}.

\section{Experiments}
\label{sec:experiments}

\begin{table*}
  \begin{center}
    \begin{tabular}{lrrrrrrrr}
      \toprule
      \multirow{2}{*}{Method}
        & \multicolumn{4}{c}{\hspace{1cm}Polyhaven} & \multicolumn{4}{c}{\hspace{1cm}SMPL} \\
        & \hspace{1cm}$\text{EPE}_{\text{all}}$ & $\text{EPE}_{\text{noc}}$ &  $>4\text{px}_{\text{all}}$ & $\text{D1}_{\text{all}}$
        & \hspace{1cm}$\text{EPE}_{\text{all}}$ & $\text{EPE}_{\text{noc}}$ &  $>4\text{px}_{\text{all}}$ & $\text{D1}_{\text{all}}$ \\
      \midrule
      CascadeStereo \cite{gu2020cascade}\textsuperscript{\textdagger} &
        $16.97$ & $14.37$ & $31.1$ & $6.77$ & $ 8.31$ & $ 6.51$ & $13.8$ & $2.97$ \\
      CFNet \cite{shen2021cfnet}\textsuperscript{\textdagger} &
        $14.50$ & $11.98$ & $31.4$ & $7.80$ & $13.28$ & $12.48$ & $ 9.8$ & $3.74$ \\
      CoExNet \cite{bangunharcana2021correlate}* &
        $ 9.78$ & $ 8.57$ & $25.9$ & $7.21$ & $ 2.98$ & $ 2.38$ & $ 8.6$ & $1.56$ \\
      FADNet++ \cite{wang2021fadnet++}* &
        $11.44$ & $10.49$ & $25.3$ & $7.82$ & $ 2.67$ & $ 1.85$ & $ 6.8$ & $1.64$ \\
      GwcNet \cite{guo2019group}\textsuperscript{\textdagger} &
        $19.97$ & $17.04$ & $35.8$ & $9.60$ & $11.27$ & $10.34$ & $14.9$ & $3.86$ \\
      IGEV-Stereo \cite{xu2023iterative}* &
        $ 5.22$ & $ 4.10$ & $16.6$ & $3.94$ & $ 1.68$ & $ 1.27$ & $ 6.2$ & $0.83$ \\
      MSNet2D \cite{shamsafar2022mobilestereonet}\textsuperscript{\textdagger} &
        $10.08$ & $ 8.69$ & $44.2$ & $5.67$ & $ 5.24$ & $ 4.44$ & $28.9$ & $2.38$ \\
      MSNet3D \cite{shamsafar2022mobilestereonet}\textsuperscript{\textdagger} &
        $14.41$ & $11.95$ & $32.3$ & $7.65$ & $ 9.78$ & $ 8.36$ & $12.3$ & $3.40$ \\
      PSMNet \cite{chang2018pyramid}\textsuperscript{\textdagger} &
        $13.19$ & $11.28$ & $37.8$ & $6.11$ & $17.38$ & $16.31$ & $17.9$ & $4.55$ \\
      \midrule
      VHS (ours)&
        $ 0.98$ & $ 0.55$ & $ 3.2$ & $0.40$ & $ 0.54$ & $ 0.41$ & $ 0.9$ & $0.10$ \\
      \bottomrule
    \end{tabular}
  \end{center}
  \caption{
    Comparison on our data
    using the model implementations from~\cite{guo2023openstereo}.
    Methods marked with * run on half resolution with inputs aligned to set minimum disparity to zero.
    \textsuperscript{\textdagger} on quarter resolution with inputs aligned to set minimum disparity to zero.
  }
  \label{tab:polyheaven_smpl}
\end{table*}

We evaluate our method in terms of average end-point error (EPE) in pixels,
proportion of errors ($>4\text{px}$ in \%) and
the D1 outlier rate~\cite{menze2015object}.
Runtime and video memory measurements follow the literature and employ automatic mixed precision.

\subsection{Benchmark Evaluation}

We first validate the correctness of our sparse-dense correlation network compared to the state-of-the-art, with 
all methods being trained on SceneFlow.
\Cref{tab:scene_flow_results} shows that our method performs competitively in terms of EPE for disparities within the range that all methods can handle. Specifically, for pixels with true disparities less than or equal to $192$ ($\text{EPE}_{\leq 192}$), our method matches with FADNet++ \cite{wang2021fadnet++}, with only three methods achieving better scores.
Notably, when evaluated on all pixels ($\text{EPE}_{\text{all}}$), our method surpasses all baseline models as we do not have any upper limit to the possible disparity.
 
\begin{figure}[tb]

\graphicspath{{figures/performance_tradeoff/}}
{
  \def\svgwidth{0.99\linewidth}
  \footnotesize
  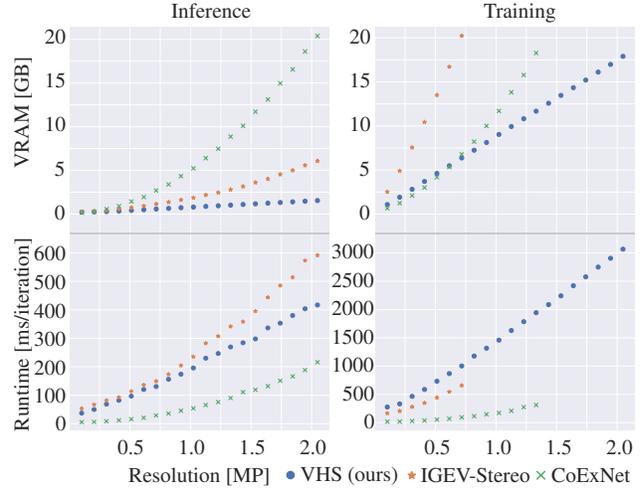
}

\caption{
  Memory and runtime statistics of our method compared to the best-performing (IGEV) and fastest (CoExNet) baseline methods.
  We fix the image height at $320$ px and increase the width, adjusting the maximum disparity to $\frac{1}{4}$ of the latter.
}
\label{fig:performance_tradeoff}
\end{figure}

Also, our method requires less memory during both inference and training as shown in \cref{fig:performance_tradeoff} and is as fast as IGEV-Stereo \cite{xu2023iterative} during inference while having a minor runtime overhead during training.

\begin{table}
  \begin{center}
    \begin{tabular}{lrrr}
      \toprule
      Method & \#Params & $\text{EPE}_{\leq 192}$ & $\text{EPE}_{\text{all}}$ \\
      \midrule
      CascadeStereo \cite{gu2020cascade}              & $10.5$M & $0.67$ & $3.30$ \\
      CFNet \cite{shen2021cfnet}                      & $23.0$M & $0.96$ & $3.06$ \\
      CoExNet \cite{bangunharcana2021correlate}       & $ 3.5$M & $0.69$ & $3.36$ \\
      FADNet++ \cite{wang2021fadnet++}                & $12.4$M & $0.88$ & $3.55$ \\ %
      GwcNet \cite{guo2019group}                      & $ 6.9$M & $0.76$ & $3.52$ \\
      IGEV-Stereo \cite{xu2023iterative}              & $12.6$M & $0.48$ & $3.01$ \\
      MSNet2D \cite{shamsafar2022mobilestereonet}     & $ 2.3$M & $1.11$ & $3.76$ \\
      MSNet3D \cite{shamsafar2022mobilestereonet}     & $ 1.8$M & $0.79$ & $3.44$ \\
      PSMNet \cite{chang2018pyramid}                  & $ 5.2$M & $1.02$ & $3.69$ \\
      \midrule
      VHS (ours)                                             & $12.7$M & $0.89$ & $2.33$ \\
      \bottomrule
    \end{tabular}
  \end{center}
  \caption{
    Comparison on SceneFlow final pass test set
    using the model implementations from~\cite{guo2023openstereo}.
  }
  \label{tab:scene_flow_results}
\end{table}

\subsection{Visual Hull Guidance}
\begin{figure*}[t]

\begin{minipage}[b]{0.19\linewidth}
  \centering
  \centerline{\includegraphics[width=\linewidth,trim={30cm 18cm 30cm 33.89cm},clip]{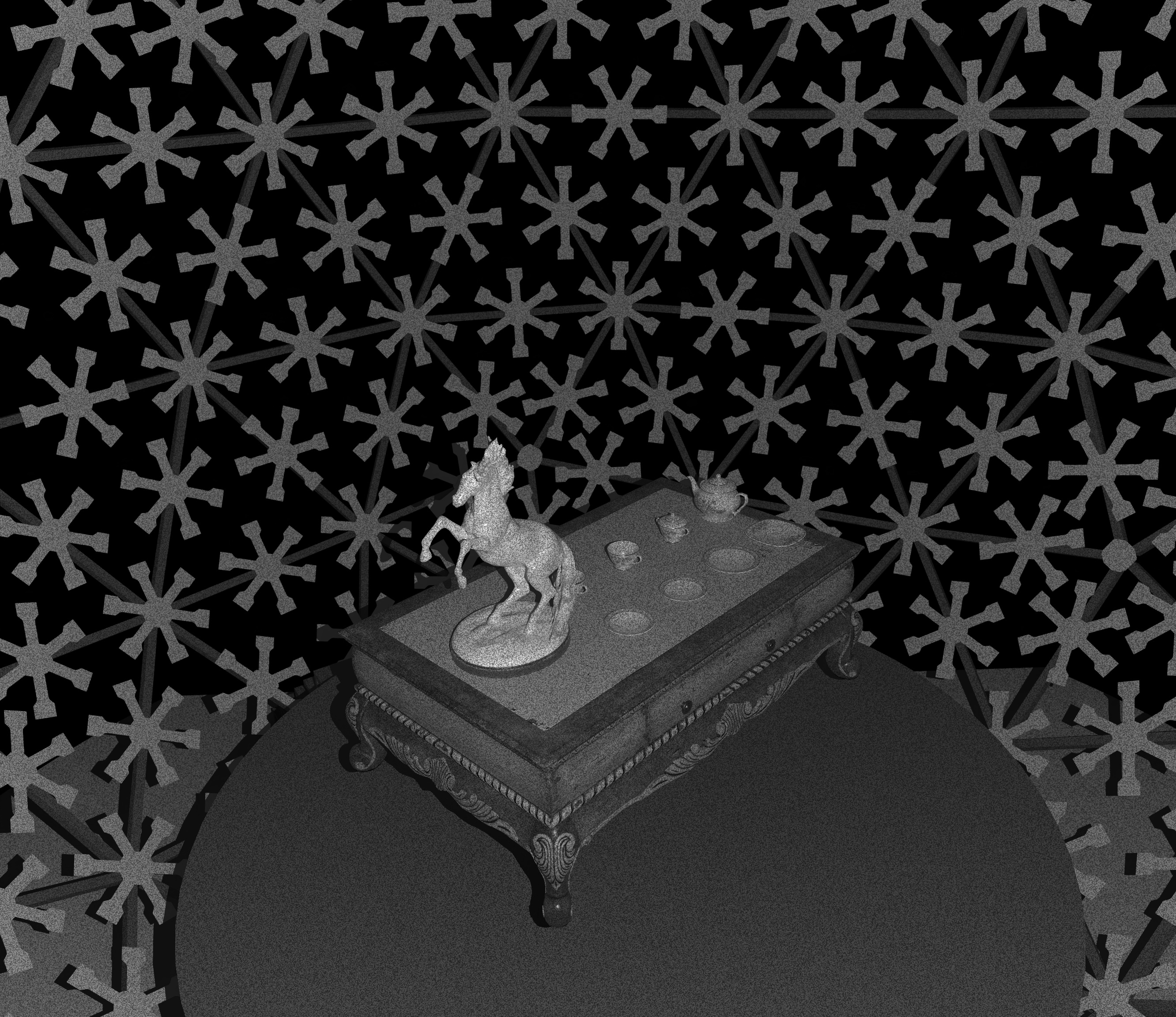}}
  \centerline{\includegraphics[width=\linewidth,trim={38cm 32cm 22cm 19.89cm},clip]{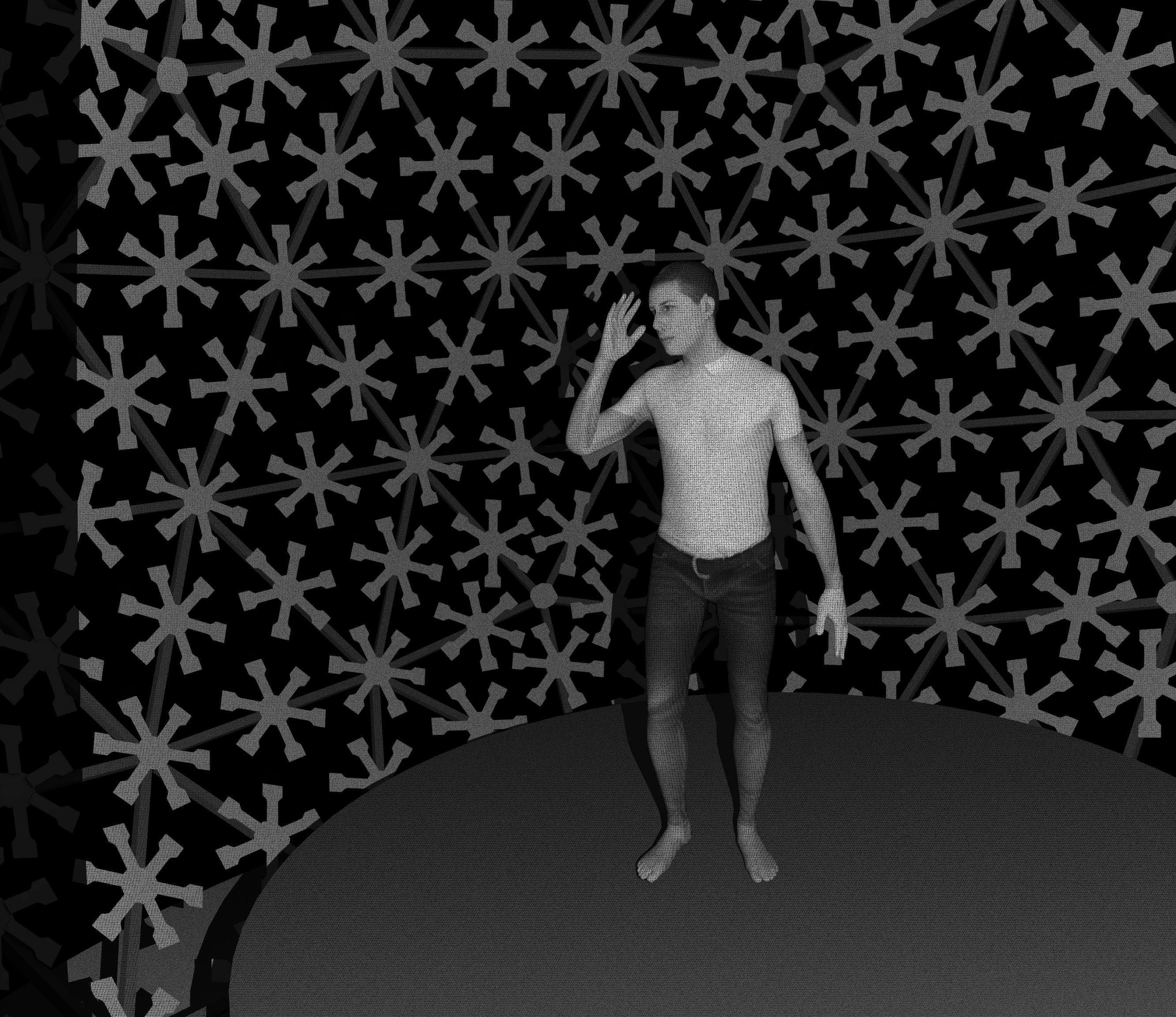}}
  \centerline{Reference (Cropped)}
  \centerline{\small Full $4608\times5328$}\medskip
\end{minipage}
\hfill
\begin{minipage}[b]{0.19\linewidth}
  \centering
  \centerline{\includegraphics[width=\linewidth,trim={30cm 18cm 30cm 33.89cm},clip]{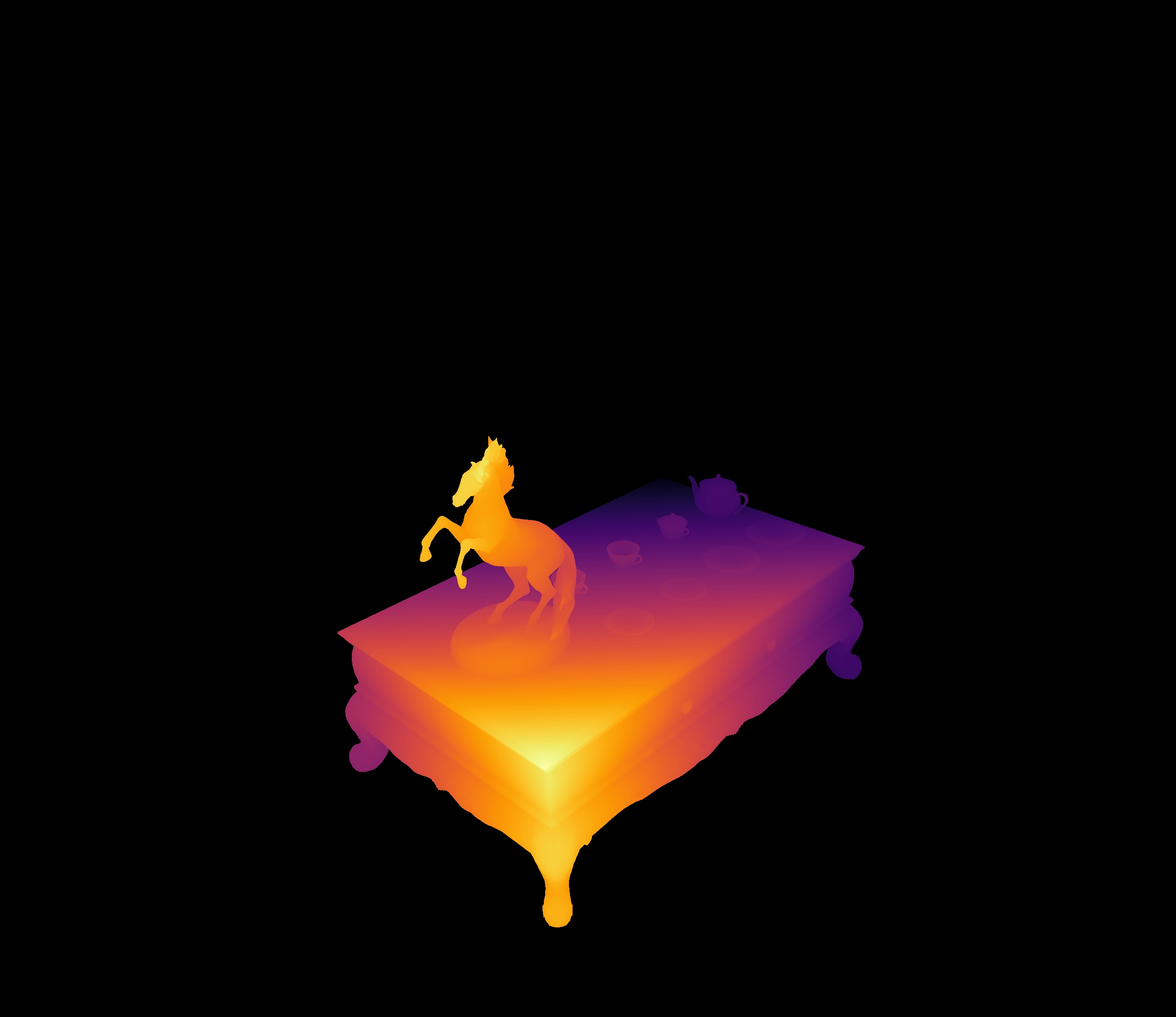}}
  \centerline{\includegraphics[width=\linewidth,trim={38cm 32cm 22cm 19.89cm},clip]{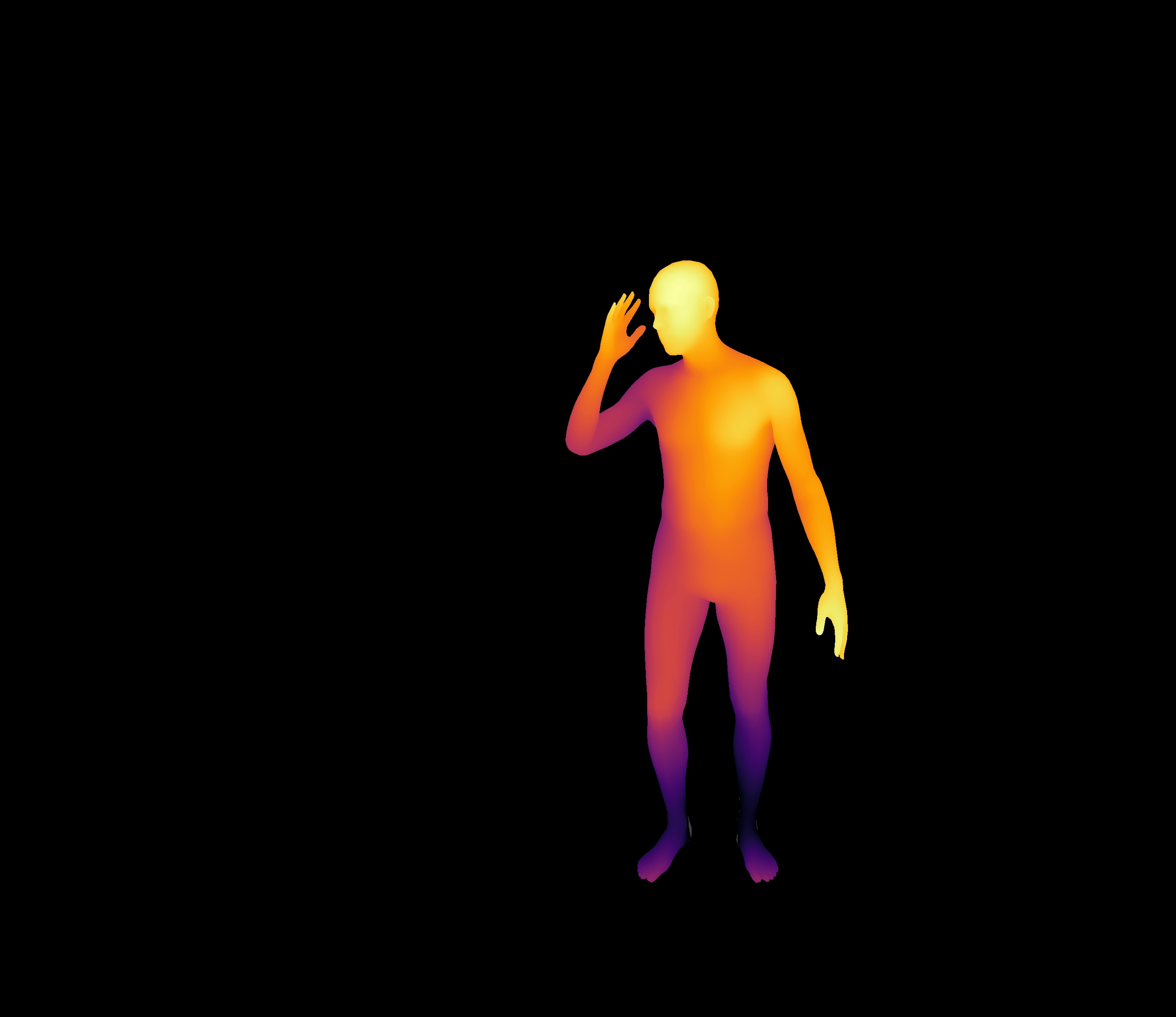}}
  \centerline{GT Disparity}
  \centerline{\small $445-721$ / $596-698$}\medskip
\end{minipage}
\hfill
\begin{minipage}[b]{0.19\linewidth}
  \centering
  \centerline{\includegraphics[width=\linewidth,trim={30cm 18cm 30cm 33.89cm},clip]{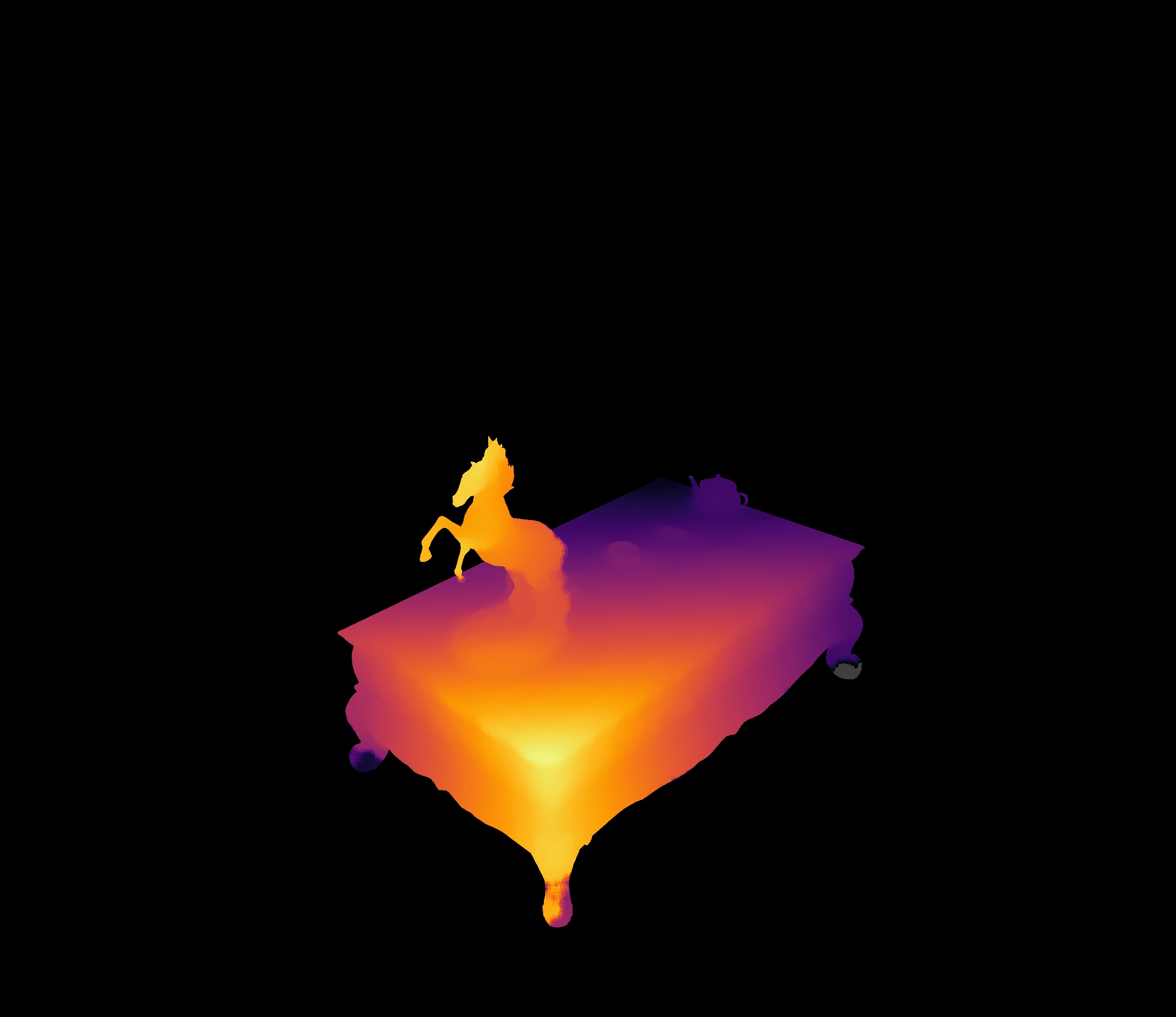}}
  \centerline{\includegraphics[width=\linewidth,trim={38cm 32cm 22cm 19.89cm},clip]{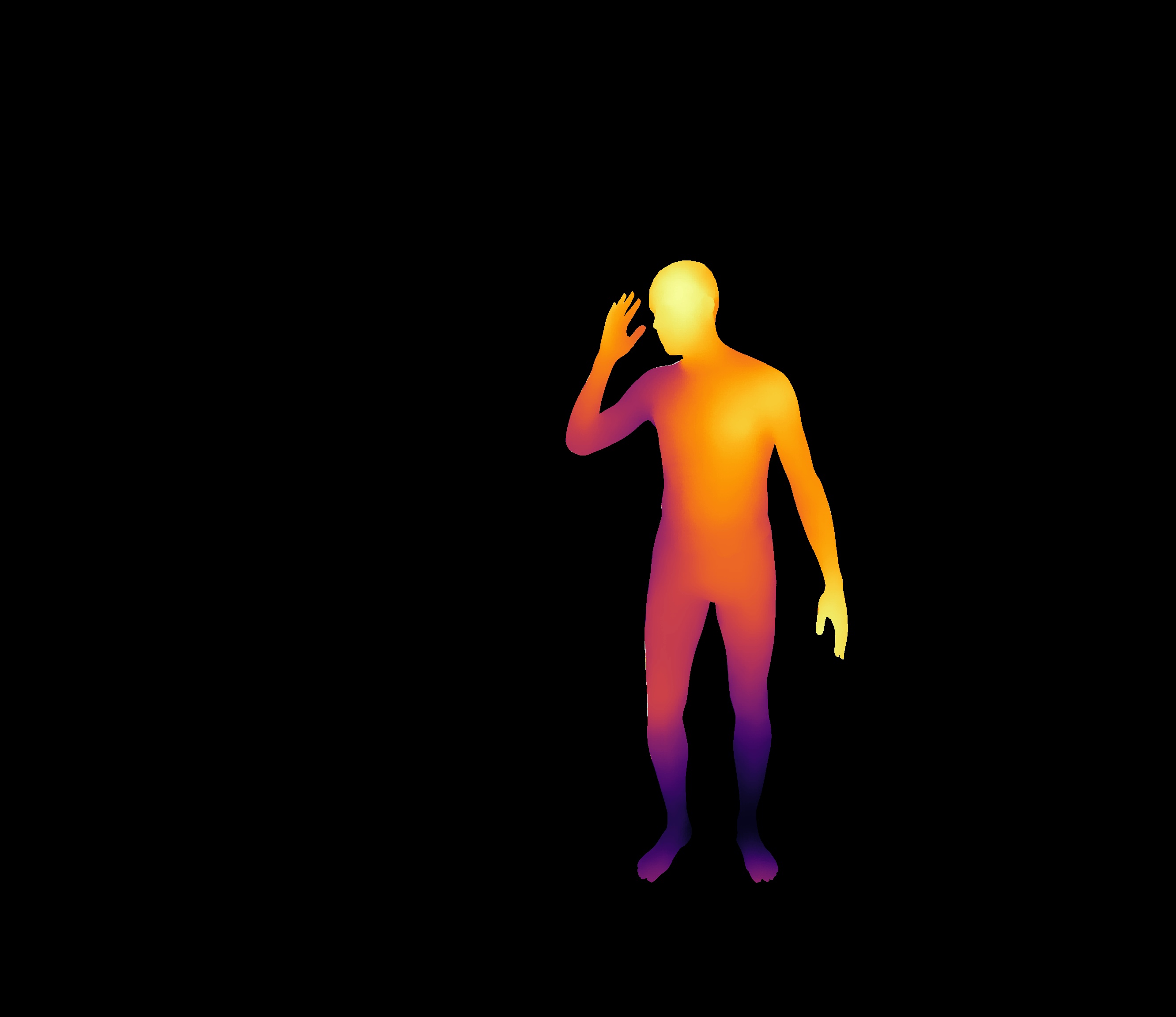}}
  \centerline{CoExNet \cite{bangunharcana2021correlate}}
  \centerline{\small EPE: $3.62$ / $1.65$}\medskip
\end{minipage}
\hfill
\begin{minipage}[b]{0.19\linewidth}
  \centering
  \centerline{\includegraphics[width=\linewidth,trim={30cm 18cm 30cm 33.89cm},clip]{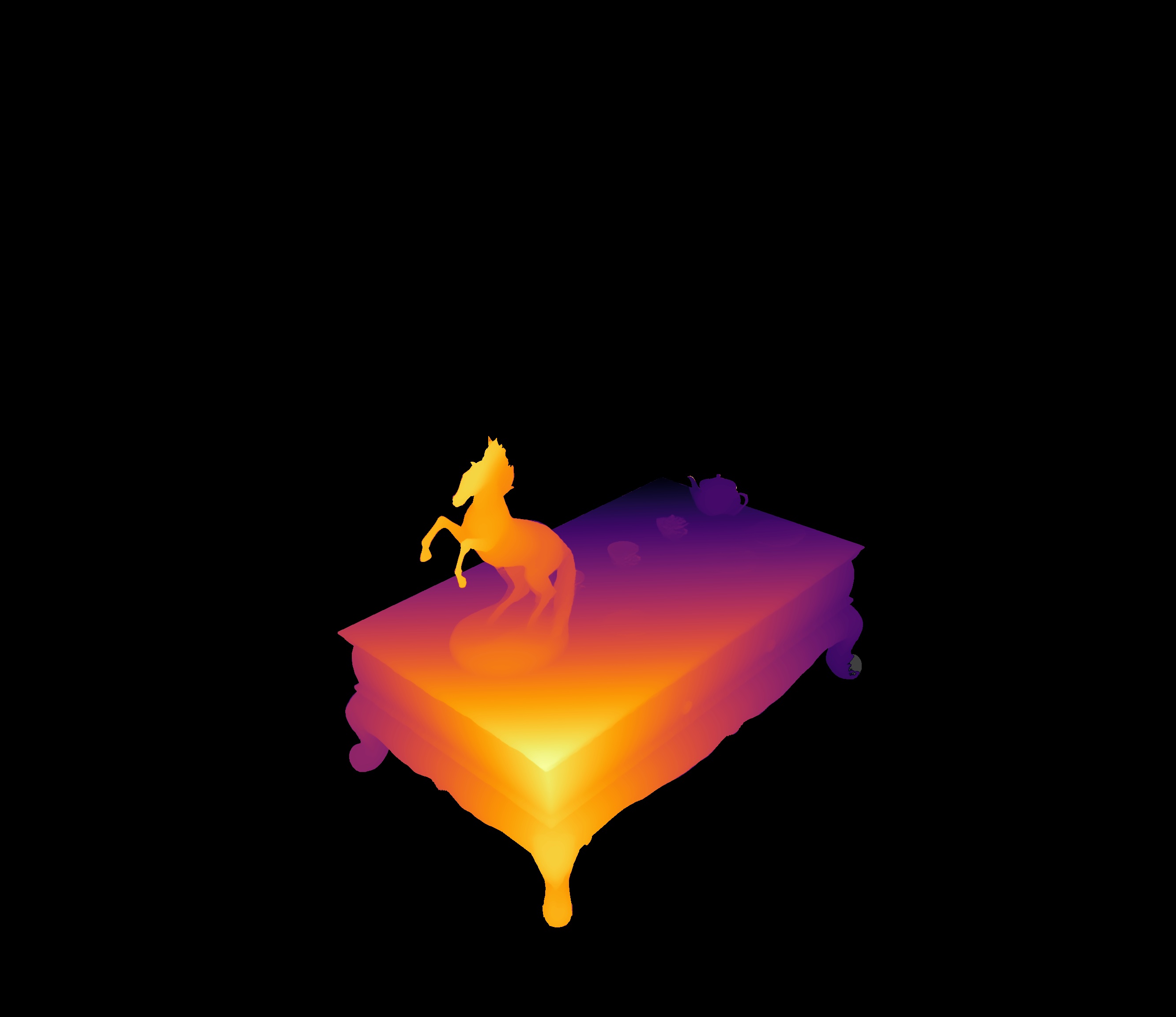}}
  \centerline{\includegraphics[width=\linewidth,trim={38cm 32cm 22cm 19.89cm},clip]{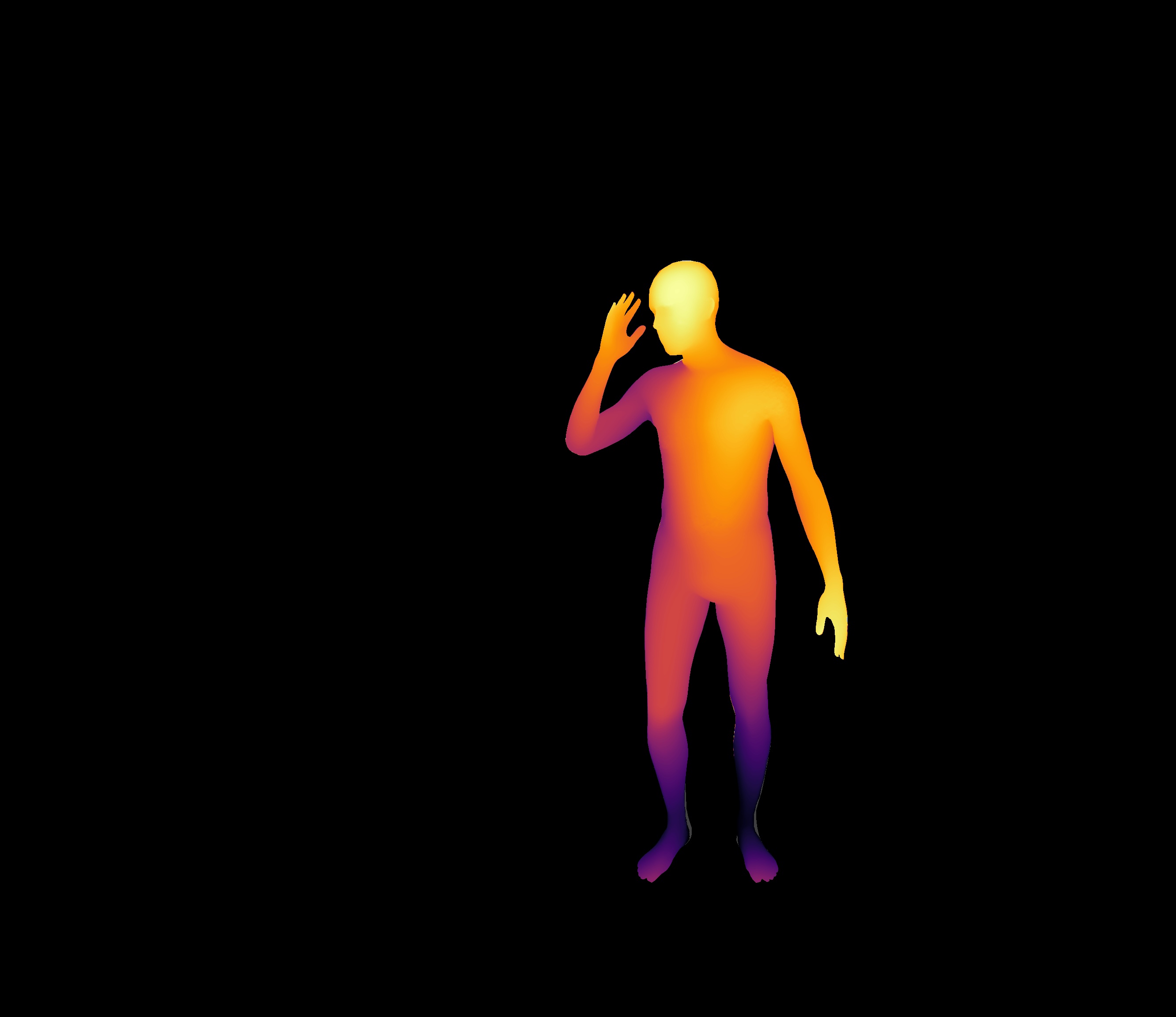}}
  \centerline{IGEV-Stereo \cite{xu2023iterative}}
  \centerline{\small EPE: $1.14$ / $0.92$}\medskip
\end{minipage}
\hfill
\begin{minipage}[b]{0.19\linewidth}
  \centering
  \centerline{\includegraphics[width=\linewidth,trim={30cm 18cm 30cm 33.89cm},clip]{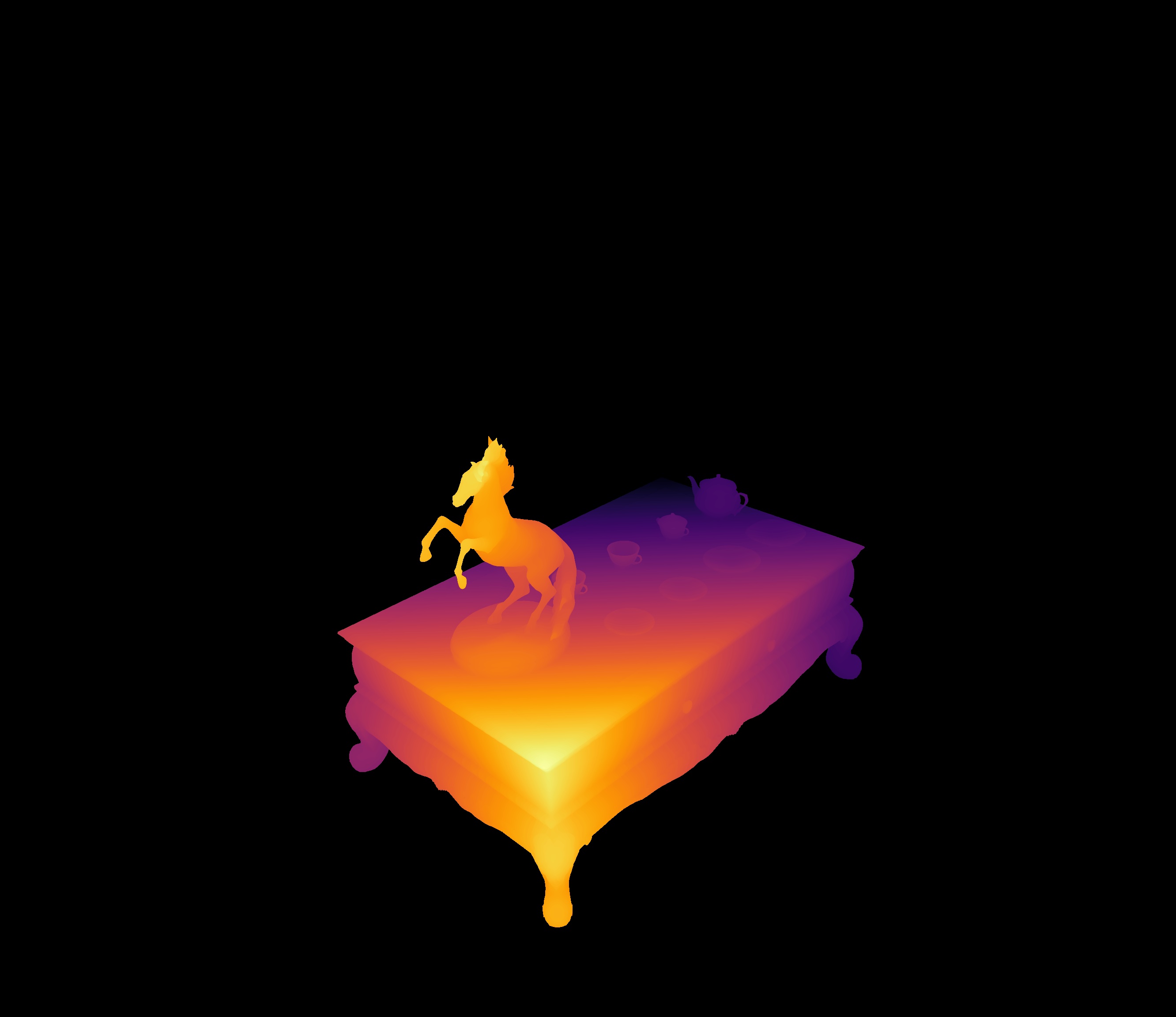}}
  \centerline{\includegraphics[width=\linewidth,trim={38cm 32cm 22cm 19.89cm},clip]{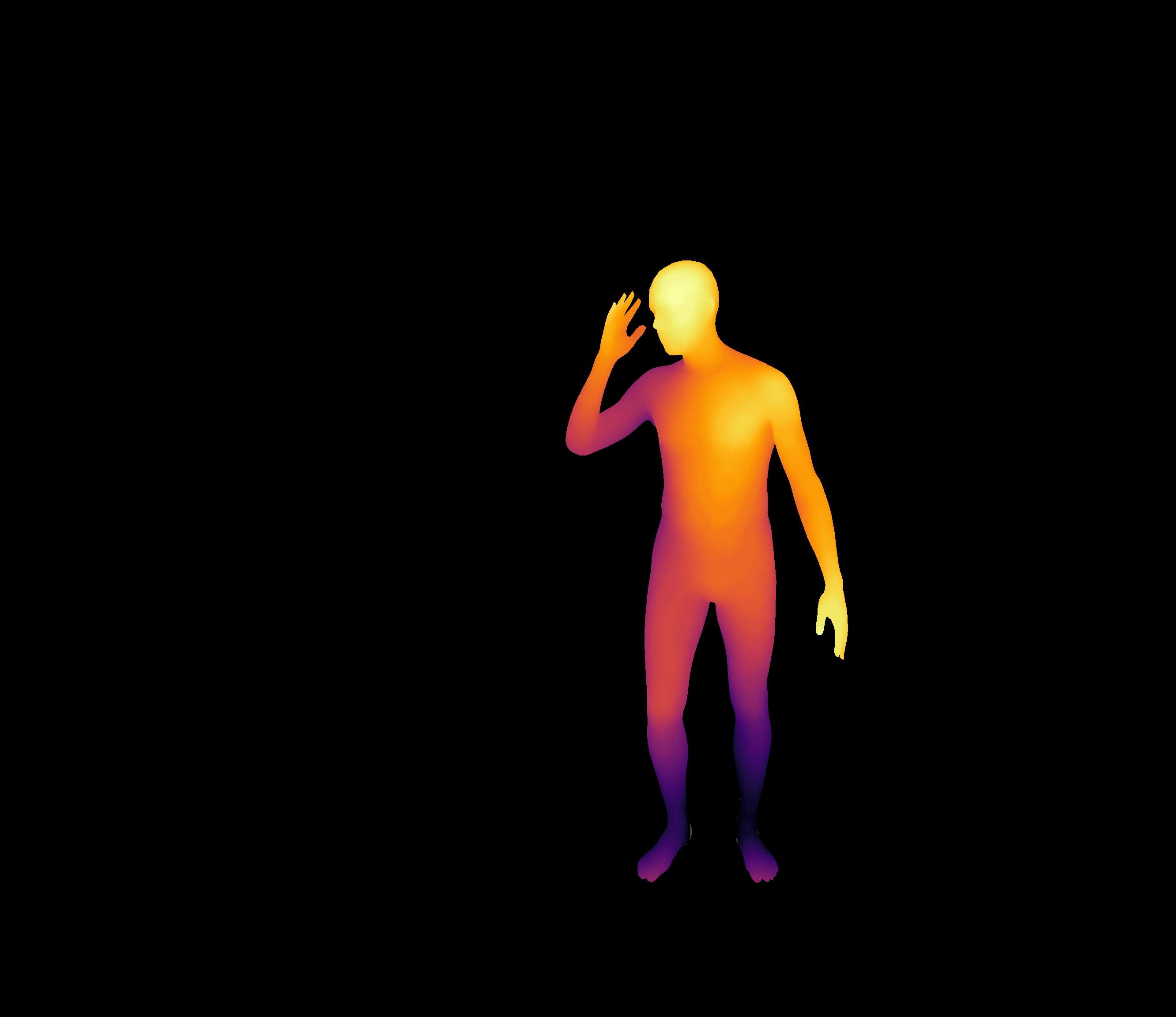}}
  \centerline{VHS (ours)}
  \centerline{\small EPE: $0.24$ / $0.42$}\medskip
\end{minipage}

\caption{
  Qualitative results on samples from the Polyhaven and SMPL test sets.
  Note the faithful reconstruction of the plates (top) and the chest (bottom) produced by our method.
  We show the range of disparity values below the GT disparity and the EPE below the methods.
}
\label{fig:qualitative}
\end{figure*}

To further demonstrate our performance on high-resolution data with larger disparities using the additional visual hull input, we evaluate our method on the two test datasets after fine tuning on the training dataset as described in \cref{sec:dataset_preparation}.
As shown in \cref{tab:polyheaven_smpl}, our method outperforms all other methods on both the Polyhaven and SMPL datasets across all metrics.
Specifically, we achieve significantly lower $\text{EPE}_{\text{all}}$ and $\text{EPE}_{\text{noc}}$ 
which indicates higher overall accuracy, and a higher accuracy  in non-occluded regions. We further highlight the robustness of our method by showing the lowest percentage of pixels with large disparity errors ($>4\text{px}_{\text{all}}$, $\text{D1}_{\text{all}}$).
We present qualitative results in \cref{fig:qualitative}.
Note that most baseline models cannot perform inference on the full resolution inputs using common hardware as they exceed the available memory (24 GB in our case) and cannot capture the large disparity values in our data as the correlation volumes are typically limited to $192$ pixels.
For this evaluation, we resort to running the models on $2\times$ or $4\times$ downsampled input images and reduce the offsets by aligning them using the known minimum ground-truth disparity of the foreground, selecting the best variant of both resolutions based on the smallest EPE.

\begin{table}
  \begin{center}
    \begin{tabular}{lrrrr}
      \toprule
      Prior & $\text{EPE}_{\text{all}}$ & $\text{EPE}_{\text{noc}}$ & $>4\text{px}_{\text{all}}$ & $\text{D1}_{\text{all}}$ \\
      \midrule
      No        & $1.48$ & $0.83$ & $4.6$ & $0.93$ \\
      Initial   & $1.29$ & $0.75$ & $4.3$ & $0.68$ \\
      Update    & $1.04$ & $0.57$ & $3.3$ & $0.46$ \\
      Both      & $0.98$ & $0.55$ & $3.2$ & $0.40$ \\
      \bottomrule
    \end{tabular}
  \end{center}
  \caption{
    Ablation of the visual hull guidance on the Polyhaven Test set.
  }
  \label{tab:ablation_visual_hull}
  \vspace{-12pt}
\end{table}

To study the performance benefit of the visual hull, we perform an ablation study on the Polyhaven test set, as shown in \cref{tab:ablation_visual_hull}. 
While applying visual hull guidance only for the initial disparity calculation already shows a minor improvement across all metrics compared to an uninformed run, the weak prior during the iterative updates yields a major gain.
Ultimately, we achieved the best results by employing visual hull guidance in both phases.
The improvement is particularly remarkable considering that the majority of the object points do not lie directly on the visual hull.

\graphicspath{{figures/mask_accuracy}}

\begin{figure}[tb]

\def\svgwidth{200pt}
\centering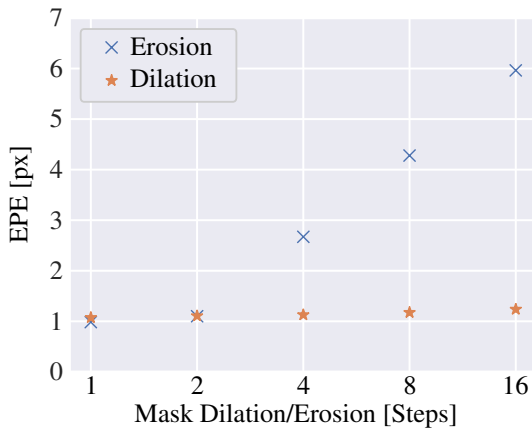

\caption{Correlation between mask accuracy and EPE, demonstrating the method's robustness to binary dilations to the correct mask.}
\label{fig:mask_accuracy}
\end{figure}

As the quality of the visual hull depends on the correctness of the masks, we additionally study the influence of incorrect matting on the performance of our method in \cref{fig:mask_accuracy}. 
We find that our method is robust against binary dilation on the masks, while larger binary erosion 
reduces the accuracy.
Intuitively, this makes sense as a correct visual hull always encloses the true surface, which is also the case for ``inflated'' visual hulls from dilated masks, while ``deflated'' hulls from eroded masks violate this assumption.

\subsection{Training Scheme}

To evaluate the impact of the memory-efficient training scheme on memory usage and runtime, we estimated these metrics for different numbers of connected updates before backpropagation in relation to the standard training procedure.
We compared a setting with full backpropagation to a setting with the detached feature extraction and measured for the former a reduction in memory usage at the cost of increased runtime for a smaller number of connected updates, as shown in \cref{tab:training_scheme}.
In comparison, the detached features offer a stable runtime even at as few as two connected layers with an even further reduction in memory usage compared to full backpropagation.

\begin{table}
  \begin{center}
    \begin{tabular}{r|>{\raggedleft\arraybackslash}p{1cm} >{\raggedleft\arraybackslash}p{1cm}|>{\raggedleft\arraybackslash}p{1cm} >{\raggedleft\arraybackslash}p{1cm}}
      \toprule
      Variant & \multicolumn{2}{c}{Full Backprop.} & \multicolumn{2}{c}{Detached Features} \\
           & GB & ms &GB & ms \\
      \midrule
      -  &$14.18$ & $377$ & - & - \\
      16 & $8.71$ & $441$ & $8.49$ & $586$ \\
      8  & $5.85$ & $497$ & $5.62$ & $584$ \\
      4  & $4.42$ & $611$ & $4.19$ & $583$ \\
      2  & $3.69$ & $840$ & $3.46$ & $583$ \\
      \bottomrule
    \end{tabular}
  \end{center}
  \caption{
    Peak memory and average runtime per iteration comparing the standard training procedure (first row) with our proposed memory-efficient training running backpropagation through the full network each time (left) and accumulating the feature gradients first (right) for different numbers of connected updates.
    Measured for a single stereo pair at $512 \times 1024$.
  }
  \label{tab:training_scheme}
\end{table}

Finally, we evaluate the impact of including pre-training on SceneFlow in our training procedure.
A network trained using only our Objaverse-XL-based dataset yields an EPE of $1.33$ on the Polyhaven test set, compared to $0.98$ of a full training, indicating a significant benefit of the hybrid approach.

\section{Conclusion}
\label{sec:conclusion}

We have presented a technique to induce visual hull priors into recurrent stereo networks to improve matching performance.
Combined with a novel sparse-dense correlation handling, our approach accurately regresses disparity for high-resolution images while retaining a favorable memory footprint and without an upper limit on the achievable disparity.

\section*{Acknowledgements}

This work has been funded by the Ministry of Culture and Science North Rhine-Westphalia under grant number PB22-063A (InVirtuo 4.0: Experimental Research in Virtual Environments),
and by the state of North Rhine-Westphalia as part of the Excellency Start-up Center.NRW (U-BO-GROW) under grant number 03ESCNW18B.
Leif Van Holland acknowledges the support of the German Research Foundation (DFG) grant KL 1142/11-2 (DFG Research Unit FOR 2535 Anticipating Human Behavior).

{\small
\bibliographystyle{ieee_fullname}
\bibliography{main}
}

\end{document}